\documentclass[10pt,twocolumn,letterpaper]{article}

\usepackage{cvpr}              %

\usepackage[pagebackref,breaklinks,colorlinks]{hyperref}

\usepackage{graphicx}
\usepackage{amsmath}
\usepackage{amssymb}
\usepackage{booktabs}
\usepackage{multirow, makecell}
\usepackage{graphicx}
\usepackage{adjustbox}
\usepackage[accsupp]{axessibility}

\usepackage{algorithm}
\usepackage{algpseudocode}

\usepackage[capitalize]{cleveref}
\crefname{section}{Sec.}{Secs.}
\Crefname{section}{Section}{Sections}
\Crefname{table}{Table}{Tables}
\crefname{table}{Tab.}{Tabs.}

\def\confName{CVPR}
\def\confYear{2024}

\usepackage{color}
\definecolor{alizarin}{rgb}{0.82, 0.1, 0.26}

\begin{document}

\title{Progressive Learning with Visual Prompt Tuning for Variable-Rate Image Compression}

\author{
Shi-Yu Qin$^{1}$
\quad
Yi-Min Zhou$^{2}$
\quad
Jin-Peng Wang$^{1}$
\quad
Bin Chen$^{2,5,6}$\thanks{Corresponding Author} \\
\quad
Bao-Yi An$^{3}$
\quad
Tao Dai$^{4}$
\quad
Shu-Tao Xia$^{1}$\\
$^{1}$Tsinghua Shenzhen International Graduate School, Tsinghua University\quad \\
$^{2}$Harbin Institute of Technology, Shenzhen
\quad\\
$^{3}$Huawei Technologies Company Ltd\quad
$^{4}$Shenzhen University \quad
$^{5}$Peng Cheng Laboratory\\
$^{6}$Guangdong Provincial Key Laboratory of Novel Security Intelligence Technologies
\\
{\tt\small qinsy23@mails.tsinghua.edu.cn, 200110126@stu.hit.edu.cn, wjp20@mails.tsinghua.edu.cn}
\\
{\tt\small chenbin2021@hit.edu.cn, anbaoyi@huawei.com, daitao.edu@gmail.com, xiast@sz.tsinghua.edu.cn}
}

\maketitle

\begin{abstract}
In this paper, we propose a progressive learning paradigm for transformer-based variable-rate image compression.
Our approach covers a wide range of compression rates with the assistance of the Layer-adaptive Prompt Module (LPM). 
Inspired by visual prompt tuning, we use LPM to extract prompts for input images and hidden features at the encoder side and decoder side, respectively, which are fed as additional information into the swin transformer layer of a pre-trained transformer-based image compression model to affect the allocation of attention region and the bits, which in turn changes the target compression ratio of the model. To ensure the network is more lightweight, we involves the integration of prompt networks with less convolutional layers. 
Exhaustive experiments show that compared to methods based on multiple models, which are optimized separately for different target rates, the proposed method arrives at the same performance with 80\% savings in parameter storage and 90\% savings in datasets. Meanwhile, our model outperforms all current variable bitrate image methods in terms of rate-distortion performance and approaches the state-of-the-art fixed bitrate image compression methods trained from scratch.
\end{abstract}
\section{Introduction}
\label{sec:intro}

\begin{figure}[t]
    \centering  
    \includegraphics[width=1\linewidth]{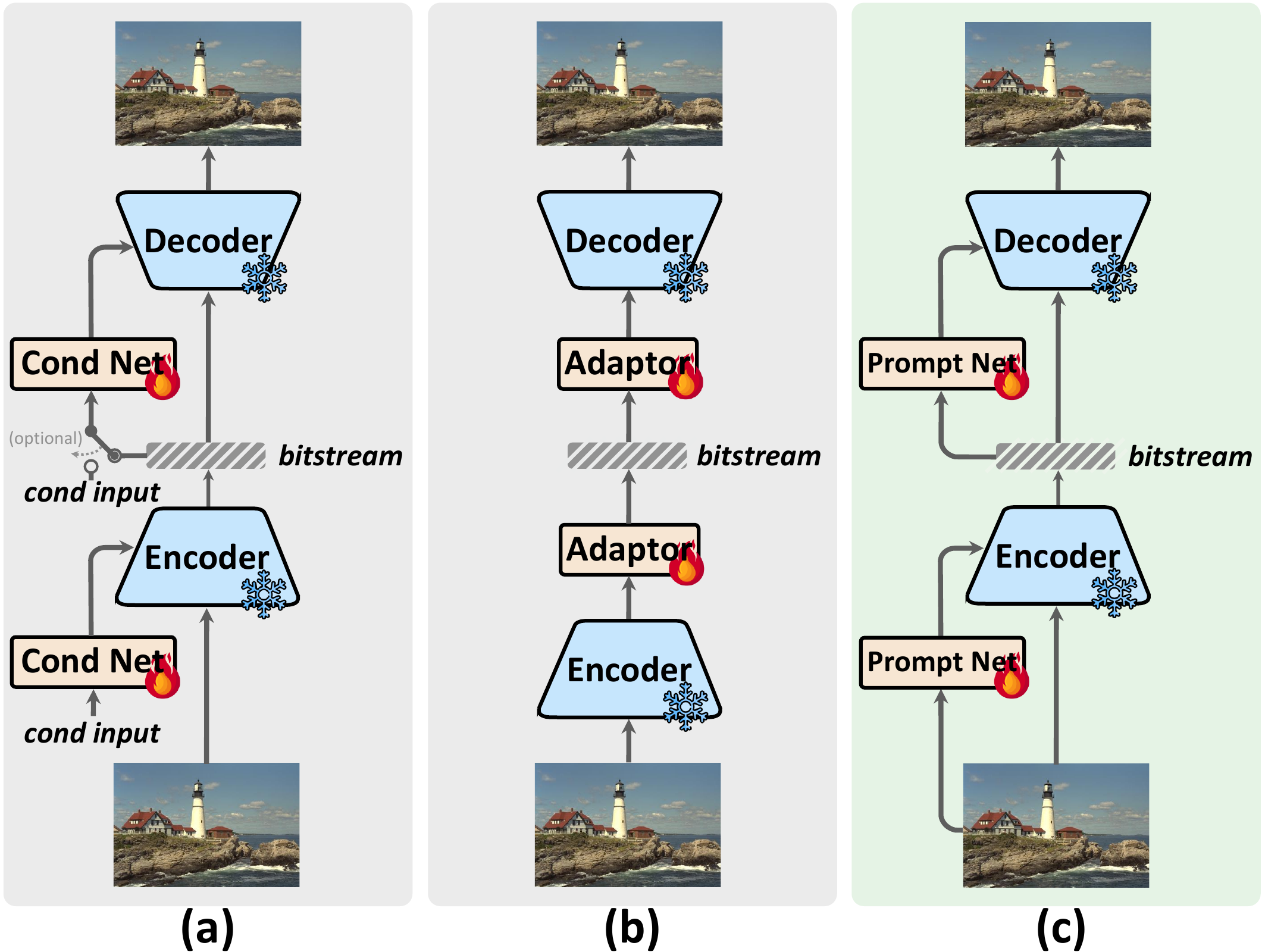}
    \caption{Comparison of existing frameworks for variable-rate methods and our proposed method. (a) Enhancing the original model by incorporating supplementary network layers. (b) Integrating external inputs alongside conditional networks for augmented functionality. (c) Our approach necessitates no extra inputs; it solely involves the integration of lightweight prompt networks.} 
    \label{intro}
\end{figure}

In the digital era, image compression plays a pivotal role in efficient data storage and transmission across various applications, ranging from web media to high-resolution medical imaging. As the demand for high-quality image communication continues to grow, the need for advanced compression techniques becomes increasingly critical. Traditionally, image compression methods, such as JPEG2K~\cite{rabbani2002overview} and VVC~\cite{bross2021overview}, have been the cornerstone in this domain. However, with the advent of deep learning, a new paradigm, known as deep image compression, has emerged, offering significant improvements over conventional techniques. Deep learning-based approaches leverage neural networks to learn data-driven, optimized compression schemes, which often result in higher compression rates with less perceptible loss in image quality.

\begin{figure*}[t]
    \centering  
    \includegraphics[width=1\linewidth]{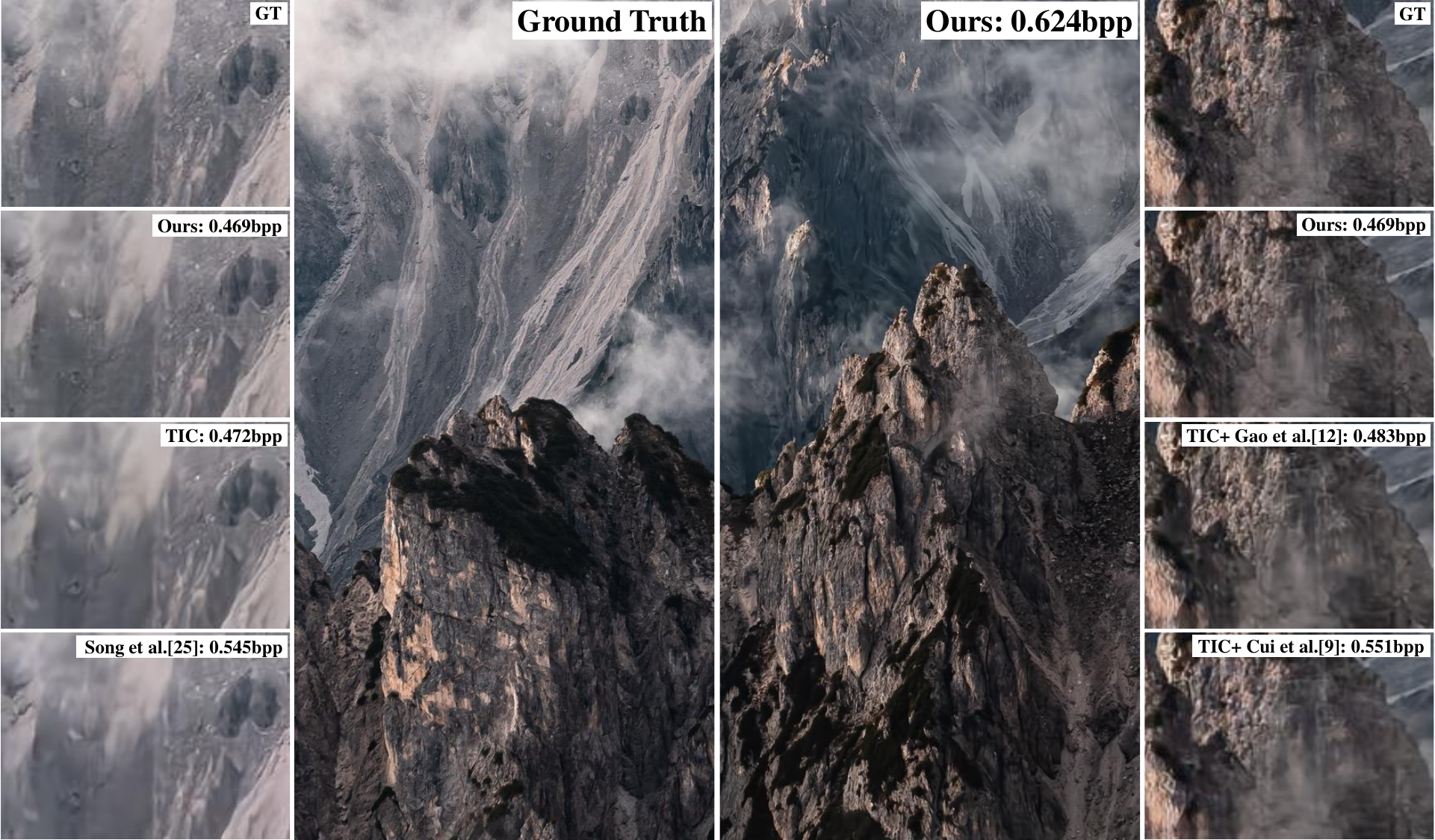}  
    \caption{Comparing our method, to the ground truth, as well as TIC~\cite{lu2021transformer} and other variable-rate methods~\cite{cui2021asymmetric,song2021variable,gao2022flexible}. It can be observed that our model produces a reconstructed image of high visual quality that closely resembles the original. The reconstructed image is sharper than the original TIC~\cite{lu2021transformer} and other variable-rate methods~\cite{cui2021asymmetric,song2021variable,gao2022flexible}, with lower bit consumption.} 
    \label{visualization}
\end{figure*}

A crucial element in deep image compression is the rate-distortion loss function, rooted in information theory~\cite{shannon1959coding}. This function balances the trade-off between the compression rate (how much data is being preserved) and the distortion (the quality loss due to compression). In traditional settings, a single hyperparameterized Lagrange multiplier $\lambda$ maps to a fixed bitrate compression model. This correspondence means that each model is trained from scratch for a specific compression rate, determined by the chosen $\lambda$ value. While this approach has its advantages of improved compression ratios, it also has significant limitations. The need to train a separate model for each target bitrate is not only time-consuming but also inefficient in the scenario of adaptive coding requirements. This implies that a substantial amount of data must be learned for each individual model, along with a considerable storage requirement to accommodate multiple compression models corresponding to different $\lambda$ values, meeting diverse compression requirements.

The introduction of variable-rate in deep image compression offers a promising solution to these challenges. By incorporating additional parameters, variable-rate methods achieve a dynamic adjustment of the target compression rate, negating the need for multiple models corresponding to different bitrates. The initial methods~\cite{toderici2015variable,toderici2017full} for variable-rate coding were based on Recurrent Neural Networks (RNNs), suffering from high computational complexity and poor rate-distortion (R-D) performance. Subsequent work mainly focused on the Variational Autoencoder (VAE) architecture. Figure \ref{intro} shows a high-level overview of different VAE-based variable-rate methods. Some approaches~\cite{cui2021asymmetric} (corresponding to Figure \ref{intro} (a)) control the target rate through external inputs and additional conditional networks, while others~\cite{choi2019variable,yang2020variable,song2021variable} (corresponding to Figure \ref{intro} (b)) have added additional layers to the existing network architecture.

These methods all improved upon CNN-based network architectures, but their adaptation to transformer-based models has not produced satisfactory outcomes. While transformer-based deep compression models have recently shown superior performance, and simply applying the aforementioned methods to these models does not fully exploit their potential. As a unique fine-tuning paradigm in transformer networks, visual prompt tuning~\cite{jia2022visual} has demonstrated significant potential in high-level tasks. Inserting a small number of additional parameters for fine-tuning can achieve or even surpass the performance of fully trained models. 

Considering the background mentioned above, this raises the question: \emph{\textbf{Can a variable bitrate fine-tuning method be designed for a transformer-based image compression model that is not only efficient in storage and sample usage, but also in time complexity?}}

As shown in Fig.~\ref{intro} (c), we introduce a progressive learning approach for variable bitrate image compression using the Transformer architecture. Our approach begins with the pretraining of a deep compression model at a designated rate. Building upon this base, we then integrate an auxiliary prompt network. Following this integration, we methodically fine-tune the rate-distortion function, gradually steering it towards the desired target rate. In both the encoding and decoding stages, we utilize a Layer-adaptive Prompt Module (LPM) to extract prompts. The LPM learns a prompt for each Swin Transformer block, and these prompts adjust the model's attention regions during fine-tuning, thereby modifying bit allocation in entropy coding. Our approach includes only storing a pre-trained backbone network model, then learning prompt networks for different compression rates to achieve adjustable target bitrates. It's noteworthy that we extensively use max-pooling layers to minimize the additional LPM, resulting in the LPM parameters comprising only 20\% of the backbone network while delivering comparable performance to independently trained models. Furthermore, our method achieves desirable results using only 10\% of the data during prompt-tuning, demonstrating sample efficiency. Moreover, compared to models separately trained for different compression rates, our approach converges faster and consumes less time. Extensive experimental results indicate that our proposed method approaches state-of-the-art fixed-rate deep image compression methods and surpasses all current variable-rate approaches.

To summarize, our contributions are as follows:
\begin{itemize}
 \item We propose a progressive learning paradigm for variable-rate image compression with transformers.
\item We designed an efficient variable-rate prompt-tuning network based on the Layer-adaptive Prompt Module, which alters the attention regions of the model and subsequently changes bit allocation.
 \item Our proposed method achieves performance comparable to state-of-the-art fixed-rate deep image compression  trained from scratch, and surpasses all existing variable-rate methods. 
\end{itemize}

\section{Related Work}
\label{sec:Related Work}

\begin{figure*}[t]
    \centering  
    \includegraphics[width=1\linewidth]{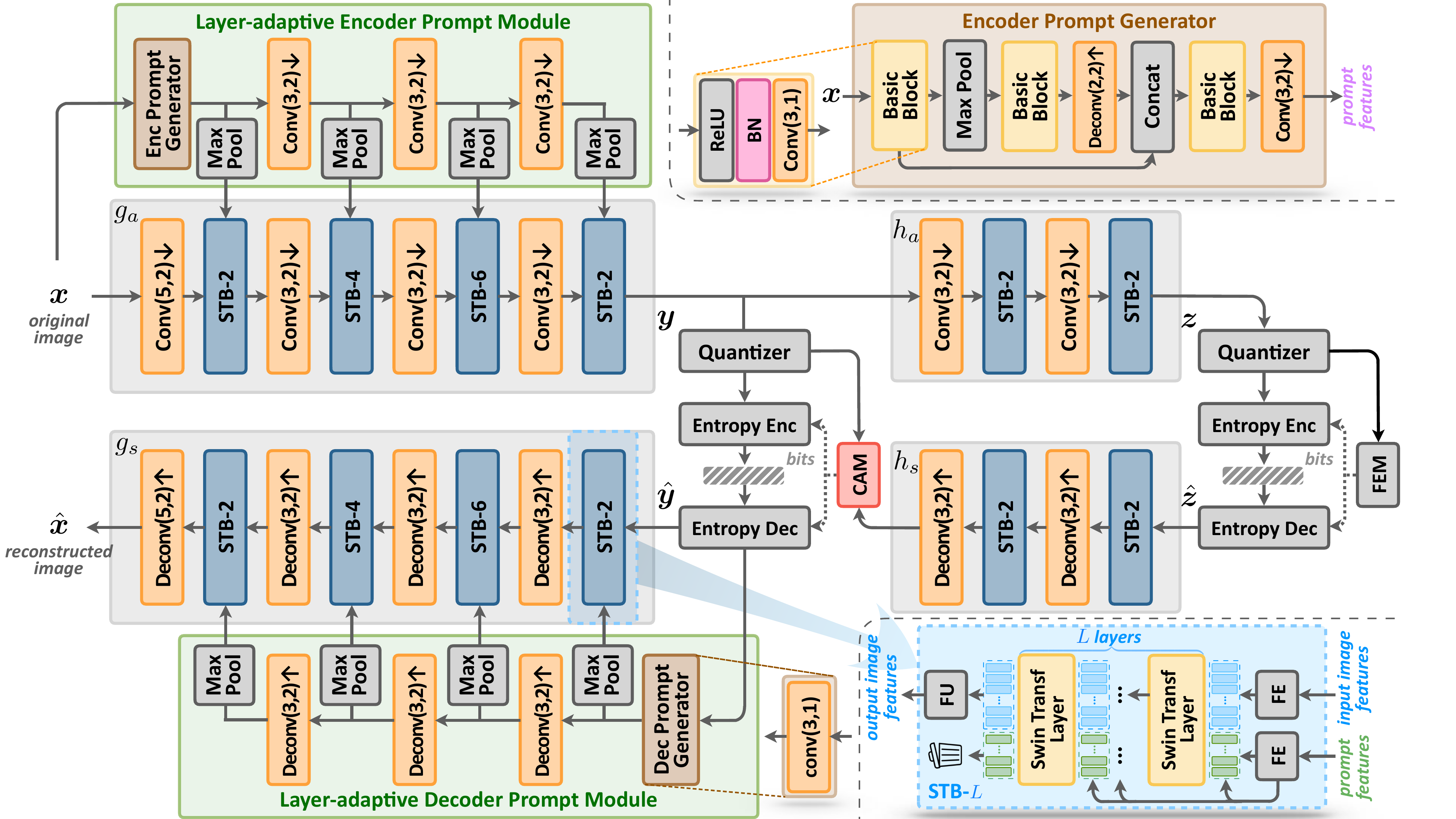}
    \caption{Our architecture. \textit{CAM} is causal attention module, \textit{FEM} is factorized entropy model. \textit{BN} is BatchNorm. \textit{conv(N,2)$\downarrow$} is a strided down convolution with $N{\times}N$ filters. \textit{deconv(N,2)$\uparrow$} is a strided up convolution with $N{\times}N$ filters. \textit{FE} and \textit{FU} respectively stand for Feature Embedding and Feature Unembedding. \textit{STB-i} represents a Swin-Transformer Block with \textit{i} layers of Swin-Transformer Layer.}
    \label{framework}
\end{figure*}

\noindent\textbf{Deep Image Compression.} Deep image compression (DIC), which uses Deep Neural Networks(DNN) to learn to minimize the distortion of a pair of source and reconstructed images while maximizing the likelihood of quantizing potential representations at low entropy coding cost (bit rate), has achieved great success with promising results. DIC typically has a trade-off between distortion and rate, which is controlled by hyper parameterized Lagrange multipliers $\lambda$ in the rate-distortion loss function, each of which corresponds to a separate model for different compression rates and qualities. The vast majority of current approaches~\cite{balle2016end,balle2018variational,cheng2020learned,minnen2018joint,lee2018context,mentzer2020high,minnen2020channel} use CNN as a basis for composing the encoder, decoder, and a priori models in the model, while some scholars~\cite{lu2021transformer,zhu2021transformer,kim2022joint,qian2022entroformer} have proposed the use of transformer-based networks due to its excellent performance on high-level visual tasks.
~\cite{lu2021transformer} introduce causal attention module (CAM) that combines the causal self attention and the multi-layer perceptron (MLP) to exploit closely-related priors for context modeling.~\cite{qian2022entroformer} used a ViT~\cite{dosovitskiy2020image} to help the entropy model capture global context information.
In our compression framework, we take advantage of the advanced transform (network structures), quantization and entropy models in the existing transformer-based methods \cite{lu2021transformer} for lossy image compression.

\noindent\textbf{Variable Rate Image Compression.} ~\cite{toderici2015variable} first implemented this concept using convolutional LSTM networks. On this foundation, ~\cite{toderici2017full} introduce a new hybrid of GRU and ResNet, which outperforms JPEK2K~\cite{rabbani2002overview} for the first time. However, since the target rate is controlled by the number of iterations, the processing time for encoding and decoding increases dramatically with the quality of the image, which makes LSTM-based methods unsuitable for real-world scenarios. 

~\cite{choi2019variable} realizes this conception to the VAE-based model with an additional autoencoder. 
However, the uncertainty in the discrete rank and bin size affects the performance of the model, while it becomes a difficulty to choose the optimal combination in the neighboring coverage areas. Moreover, the additional fully-connected layer makes the newly added network insufficiently lightweight.
~\cite{yang2020variable} proposed modulated autoencoders, where the representations of a shared autoencoder are adapted to the specific R-D tradeoff via a
modulation network. Similar to the approach of~\cite{choi2019variable}, the additional fully connected layer creates a larger storage burden.~\cite{cui2021asymmetric} proposed asymmetric gained variational autoencoder (AG-VAE), which utilizes a pair of gain units to achieve discrete rate adaptation in one single model. With channel affine transformations, the method exhibits comparable performance to independently trained models. \cite{song2021variable} proposed a versatile deep image compression network based on spatial feature transform, which takes a source image and a corresponding quality map as inputs and produce a compressed image with variable rates. \cite{gao2022flexible} proposed a coding method for NIC based on semi-amortized inference and adaptive quantization.

\noindent\textbf{Prompt Tuning.} Prompting is initially proposed in Natural Language Processing~\cite{brown2020language,liu2023pre}. The principle is to adapt downstream tasks with limited annotated data to the original pre-trained task at minimal cost, so that the pre-trained knowledge can be utilized to solve the downstream problem. In prompt engineering, their values are often chosen by heuristic, tokens are parameterized by learnable parameters and their parameters are updated via gradient descent to adapt transformers to the downstream tasks. ~\cite{brown2020language} demonstrates strong generalization to downstream transfer learning tasks even in the few-shot or zero-shot settings with manually chosen prompts in GPT3. Due to its simplicity and with the success of transformer-based visual models such as the ViT~\cite{dosovitskiy2020image}, ~\cite{jia2022visual} first extended this concept to the visual domain by adding additional embedding to the transformer layer. ~\cite{jin2023instance} proposed an instance prompt tuning approach based on this improvement. ~\cite{chen2023transtic} applied this method to the field of image compression to fine-tune downstream task accuracy, but the resulting extra computational load and parameters were so large that they even exceeded twice the size of the backbone network at the encoder end, seemingly deviating from the original intent of introducing prompt tuning.
\section{Proposed Method}

\begin{figure*}[t]
    \centering  
    \includegraphics[width=0.8\linewidth]{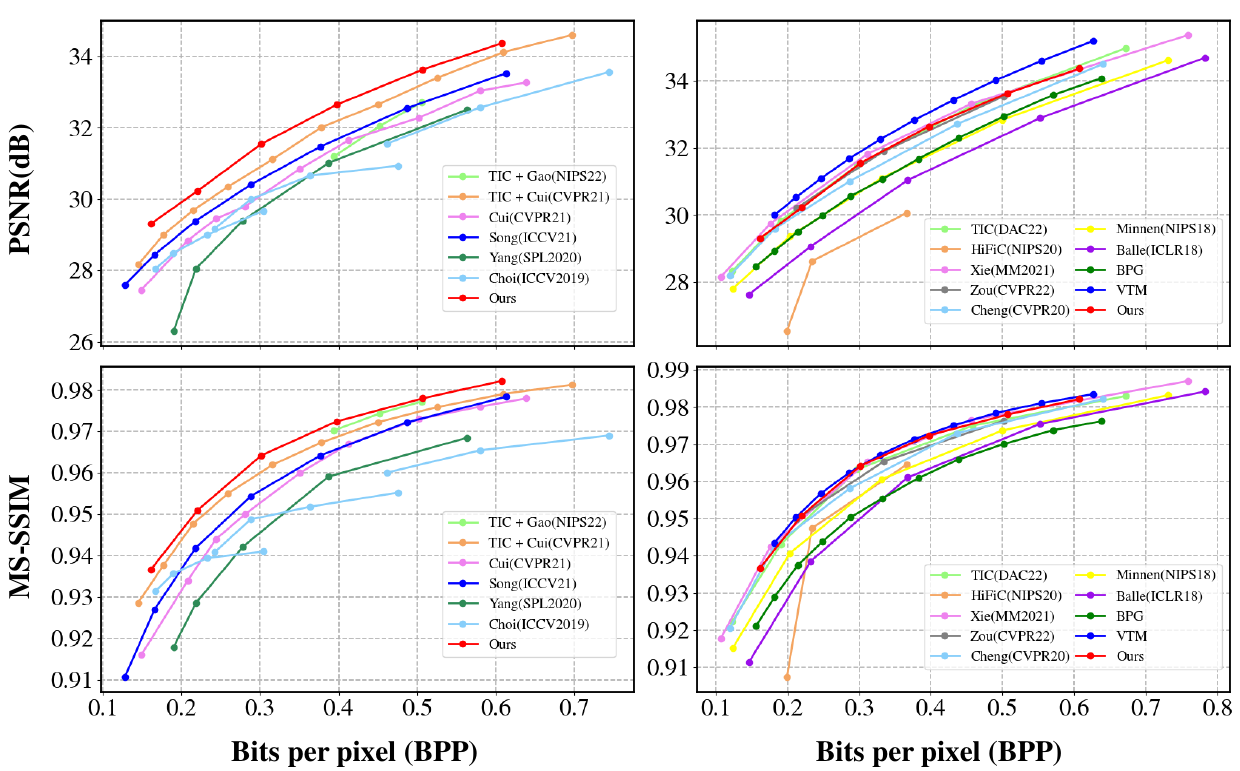}  
    \caption{PNSR and MS-SSIM comparison to baseline methods on Kodak dataset~\cite{franzen1999kodak}. The left column presents a comparison with variable-rate methods, while the right column features comparisons with fixed-rate methods.} 
    \label{main experiment Kodak}
\end{figure*}

\subsection{Overview}
Since our approach is built upon TIC~\cite{lu2021transformer}, we would briefly introduce the basic pipeline for better understanding. Figure \ref{framework} demonstrates the proposed architecture. On top of the baseline network, our method incorporates Layer-adaptive Prompt Module(LPM) for for each of the two components-encoder and decoder. We describe the detailed procedure of our network designed for image compression in the following.
 
We first encode an image $\mathbf{x}$ using an encoder $g_a(\cdot,\cdot;\phi)$, and the Layer-adaptive Encoder Prompt Module $P_E(\cdot;\psi)$ provides prompts $\mathbf{p}_e$ for the encoding process, generating a latent feature $\mathbf{y}$, as indicated below
\begin{equation} 
\mathbf{y} = g_a(\mathbf{x}, \mathbf{p}_e; \phi), ~~\text{where} ~~\mathbf{p}_e = P_E(\mathbf{x}; \psi).
\end{equation}
Following this, a hyper-encoder $h_a(\cdot;\xi)$ generates hyperprior information $\mathbf{z}$ from latent feature $\mathbf{y}$. The feature $\mathbf{z}$ captures spatial redundancies within the latent feature $\mathbf{y}$ and models the probability of $\mathbf{y}$ conditionally. Subsequently, the quantized hyperprior information $\mathbf{\hat{z}}=Q(\mathbf{z})$ is sent to hyper-decoder $h_s(\cdot;\zeta)$  to generate hyperpriors for the causal attention module, which are then utilized for the final context prediction.

\begin{equation} 
\mathbf{\hat{x}} = g_s(\mathbf{\hat{y}}, \mathbf{p}_d; \theta), ~~\text{where} ~~\mathbf{p}_d = P_D(\mathbf{\hat{y}}; \omega).
\end{equation}
In the above equation, $\phi$, $\psi$, $\xi$, $\zeta$, $\theta$ and $\omega$ represent respective network parameters.

\subsection{Layer-adaptive Prompt Module}
LPM is fundamental to the architecture's overall functionality. It is responsible for generating prompt parameters, which are crucial for fine-tuning the backbone network. The LPM consists of two main components: the Enc/Dec Prompt Generator, and a convolutional downsampling adjustment stage specifically designed to accommodate various Swin-Transformer Block (STB). 

\noindent\textbf{Encoder/Decoder Prompt Generator} On the encoder side, given the large size and substantial spatial redundancy in the original image, we deliberately designed a simple Enc Prompt Generator (EPG). As shown in Figure \ref{framework}, EPG consists of basic blocks, convolutional layers, and max-pooling layers. To enhance feature learning, we opted not to use interpolation upsampling but instead incorporated learnable deconvolutional layers. The input to EPG is the original image $\mathbf{x}\in\mathbb{R}^{H \times W \times 3}$, and the output is the initial prompt $\mathbf{p}_{e,0}\in\mathbb{R}^{\frac{H}{2} \times \frac{W}{2} \times 1}$.

On the decoder side, since entropy coding has been completed beforehand, the paramount task at this stage is to adjust the decoder for improved reconstruction of the modified latent features. Additionally, considering the lightweight nature of the supplementary network, we incorporated only a single convolutional layer as the Dec Prompt Generator (DPG). The input to DPG is the quantized latent features $\mathbf{\hat{y}}\in\mathbb{R}^{H^{'} \times W^{'} \times C^{'}}$, and the output is the initial prompt $\mathbf{p}_{d,0}\in\mathbb{R}^{H^{'} \times W^{'} \times C^{'}}$.

\noindent\textbf{Layer-adaptive Transformations} To ensure that $i$-stage of STB has a distinct prompt $\mathbf{p}_i$, we equipped each STB with a convolutional layer to learn prompts more suitable for the current block. However, for each Swin-Transformer Layer (STL) within the STB, to ensure parameter efficiency, we utilized only max-pooling layers to extract features $\mathbf{{p^{'}}}_i$ at each layer. In the encoding phase, the procedure unfolds as follows:
\begin{align}
\mathbf{p}_{e,i} &= \text{Conv}(\mathbf{p}_{e,i-1})\\
\mathbf{{p^{'}}}_{e,i}=\text{Max Pool}&(\mathbf{p}_{e,i}) ~~\text{where} ~~i=1,2,3,4 \nonumber.
\end{align}
The decoding stage follows a similar process.

\begin{figure*}[t]
    \centering  
    \includegraphics[width=0.8\linewidth]{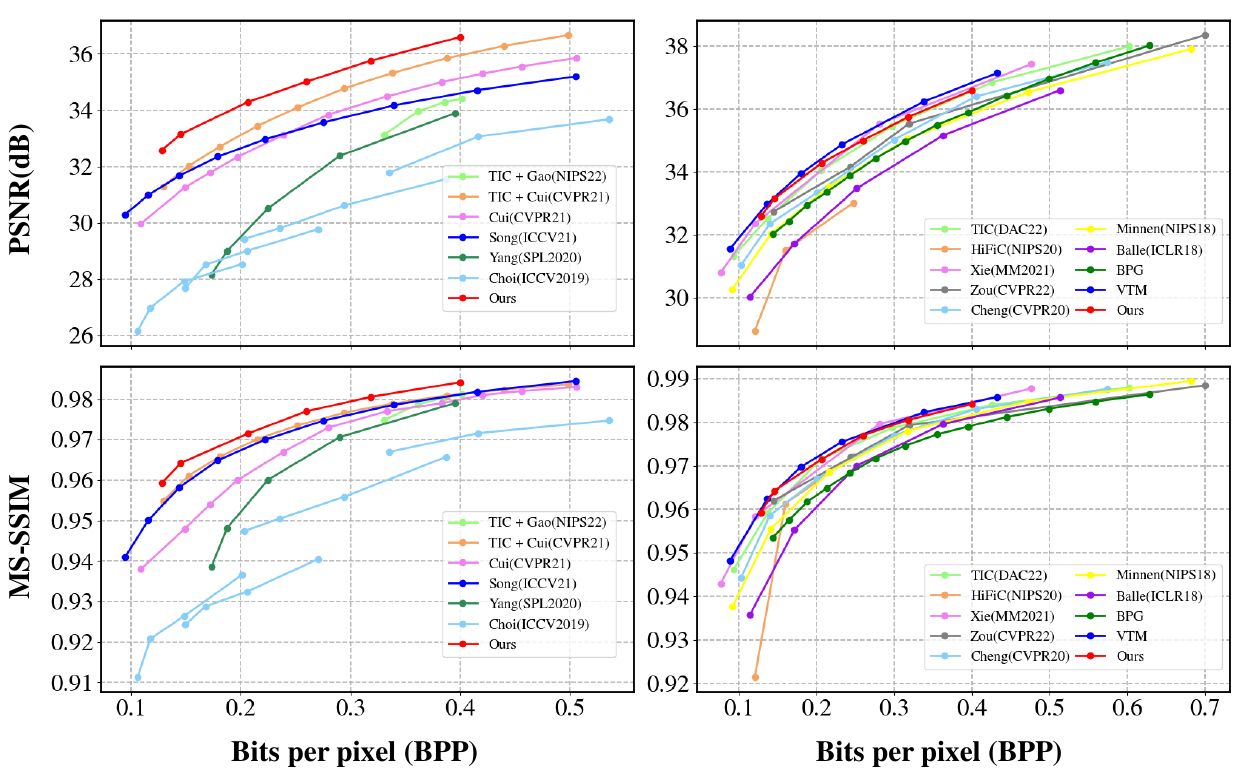}  
    \caption{PNSR and MS-SSIM comparison to baseline methods on CLIC2020 dataset~\cite{toderici2020clic}. The left column presents a comparison with variable-rate methods, while the right column features comparisons with fixed-rate methods.} 
    \label{main experiment CLIC2020}
\end{figure*}

\subsection{Swin-Transformer Block with Prompt Tuning}
STB is crucial to the transformer-based image compression model, and one significant reason is that window-based multi-head self-attention (W-MSA) and shifted window-based multi-head self-attention (SW-MSA) effectively capture the long-range correlation for better information embedding.
This step is also where the prompt comes into play. As shown in Figure \ref{framework}, STB-$i$ represents an assembly of $i$ Swin-Transformer Layers (STL). For each (S)W-MSA in an STL, it involves two inputs: image embedding $\mathbf{I}\in\mathbb{R}^{HW \times C}$ and prompt embedding $\mathbf{P}\in\mathbb{R}^{\frac{HW}{4} \times C}$. They are first reshaped into shapes $\mathbf{I}\in\mathbb{R}^{H\times W \times C}$ and $\mathbf{P}\in\mathbb{R}^{\frac{H}{2}\times \frac{W}{2} \times C}$, respectively.  Subsequently, window partitions are applied to $\mathbf{I}$ and $\mathbf{P}$ with window sizes $d$ and $\frac{d}{2}$ followed by window unfolding along the token dimension in each window to obtain $\mathbf{I}\in\mathbb{R}^{N\times d^{2} \times C}$ and $\mathbf{P}\in\mathbb{R}^{N\times \frac{d^{2}}{4} \times C}$, where $N = H \times W / d^2$. The self-attention corresponding to a specific head in a window is as shown below:
\begin{equation}
\text{Attnetion}(\mathbf{Q},\mathbf{K},\mathbf{V}) = \text{Softmax}(\mathbf{Q}\mathbf{K}^{T}/\sqrt{C} + \mathbf{B})\mathbf{V},
\end{equation}
where $\mathbf{Q}=\mathbf{I}\mathbf{W}_Q$, $\mathbf{K}=[\mathbf{I};\mathbf{P}]\mathbf{W}_K$, $\mathbf{V}=[\mathbf{I};\mathbf{P}]\mathbf{W}_V$. In the above expression, $\mathbf{I}\in\mathbb{R}^{d^{2} \times C}$ and $[\mathbf{I};\mathbf{P}]\in\mathbb{R}^{(d^{2} + \frac{d^{2}}{4}) \times C}$ respectively represent the window unfolding of image embedding and the concatenation of the unfolding of image and prompt embedding. $\mathbf{W}_Q$, $\mathbf{W}_K$, $\mathbf{W}_V\in\mathbb{R}^{C \times C}$ and $\mathbf{B}\in\mathbb{R}^{d^{2} \times (d^{2} + \frac{d^{2}}{4})}$ are frozen pre-trained parameters. It is worth noting that during the prompt tuning process, pre-trained feature $\mathbf{B^{'}}\in\mathbb{R}^{d^{2} \times d^{2}}$ is expanded to $\mathbf{B}\in\mathbb{R}^{d^{2} \times (d^{2} + \frac{d^{2}}{4})}$ by adding zero features. After that, we separate the obtained attention features $\mathbf{A}\in\mathbb{R}^{N\times (d^{2} + \frac{d^{2}}{4}) \times C}$ into $\mathbf{A}_1\in\mathbb{R}^{N\times d^{2} \times C}$ and $\mathbf{A}_2\in\mathbb{R}^{N\times \frac{d^{2}}{4} \times C}$, and perform the window reverse and resize operation yields (S)W-MSA outputs $\mathbf{I^{'}}\in\mathbb{R}^{HW \times C}$ and $\mathbf{P^{'}}\in\mathbb{R}^{\frac{HW}{4} \times C}$, while $\mathbf{P^{'}}$ will not continue to participate in the subsequent processes.

\subsection{Variable Rate-distortion Loss}

The goal of deep image compression is to minimize the rate-distortion loss function, which reflects the trade-off between the length of the bitstream and the distortion between the reconstructed image and the original image. In our paradigm, we first pre-train a model for a target rate with respect to a specific Lagrange multiplier $\lambda_0$:
\begin{eqnarray}
    \mathcal{L}_0 = R_{\tau}(Q(g_a(\mathbf{x};{\color{red}{\phi}}))) + \lambda_0 \cdot D(\mathbf{x}, g_s(Q(g_a(\mathbf{x};{\color{red}{\phi}}));{\color{red}{\theta}})),
\end{eqnarray}
where $R_{\tau(\cdot)}$ represents the expected code length (bitrate) of the quantized latent feature and $D(\cdot, \cdot)$ measures the distortion between the original image and the reconstructed image. Upon obtaining well-trained parameters $\phi$, $\xi$, $\zeta$ and $\theta$ for the backbone network, we integrate the Layer-adaptive Prompt Module $P_E(\cdot;\psi)$ and $P_D(\cdot;\omega)$ and adjust the Lagrange multiplier $\lambda_i$: 
\begin{align}
    \mathcal{L}_i =R_{\tau}(\hat{\mathbf{y}}) + \lambda_i \cdot D(\mathbf{x}, g_s(\hat{\mathbf{y}}, P_D(\hat{\mathbf{y}};{\color{red}{\omega}});{\color{blue}{\theta}}))\\
    \text{where}~~ \hat{\mathbf{y}} = Q(g_a(\mathbf{x}, P_E(\mathbf{x}; {\color{red}{\psi}});{\color{blue}{\phi}})) \nonumber.
\end{align}
After training the parameters $\psi$ and $\omega$, we obtain variable compression models tailored for the $i$-th target rate.

\section{Experiments}

\begin{figure*}[t]
    \centering  
    \includegraphics[width=1\linewidth]{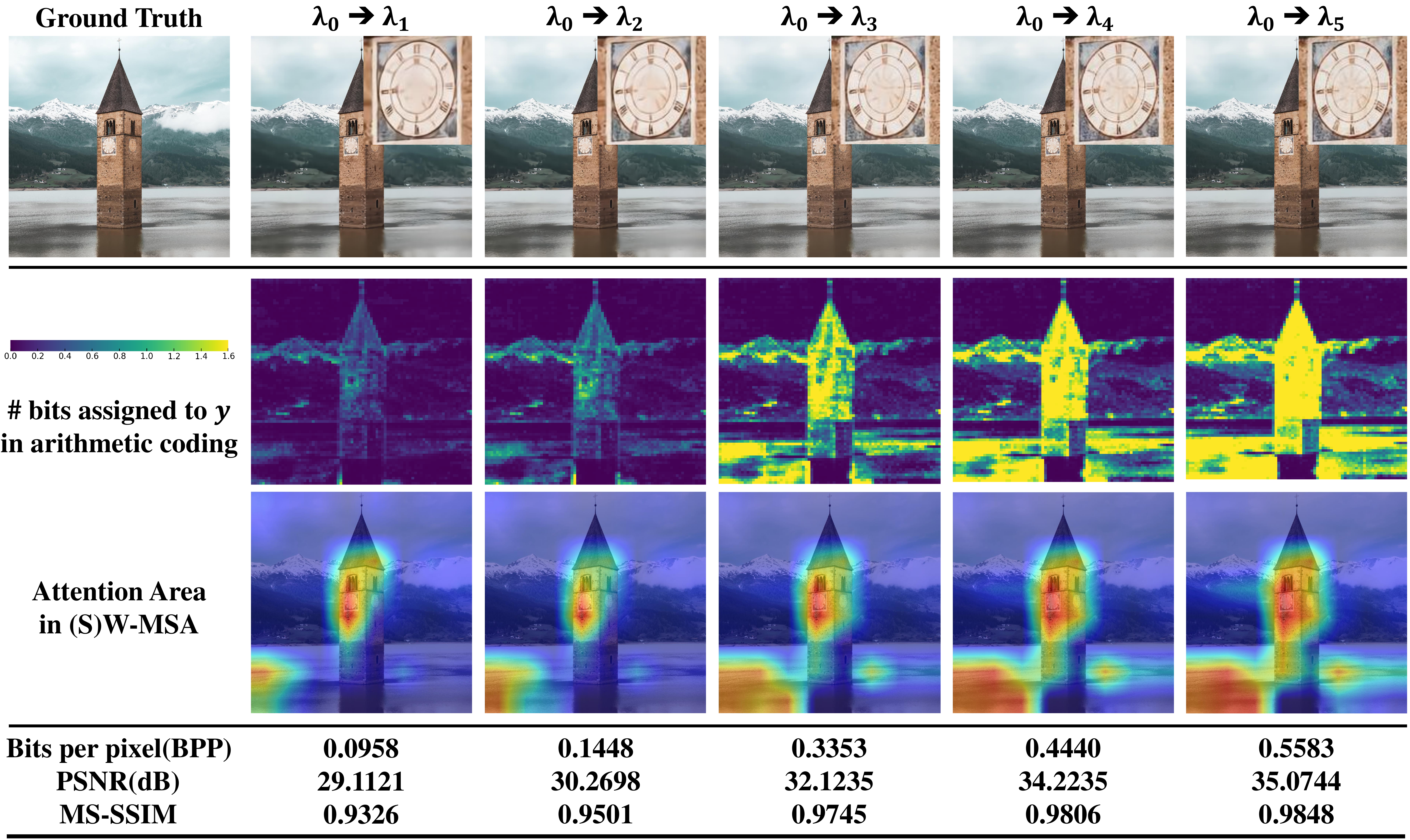}  
    \caption{Our variable-rate image compression outputs for different values $\lambda_i$, $\psi$ and $\omega$. The first row represents the reconstructed images and details shown for different compression rate models after prompt tuning. The second row presents the bit allocation map of the latent features of the compression process. The values of bit allocation maps denote the average negative log-likelihood of each element in $\hat{\mathbf{y}}$ across all channels. The third row represents the visualization results of the attention map at the encoder side after adding prompts. The fourth row demonstrates the rate and distortion measures for images with different levels of compression.} 
    \label{bitmap}
\end{figure*}

\subsection{Experimental Setup}
\noindent\textbf{Datasets.} In the training phase, we used the Flickr30k dataset~\cite{young2014image}, which has 31783 images, each of which is randomly cropped to $256\times 256$ size. We conduct experiments to show qualitative and quantitative results on both
 Kodak~\cite{franzen1999kodak} and CLIC2020~\cite{toderici2020clic}.

\noindent\textbf{Implementation details.} Our model is implemented using Pytorch. The entire training process is split into two stages. In the first stage, we set the Lagrange multiplier $\lambda_0 = 0.0067$ and solely train the parameters of the backbone network $g_a(\cdot;\phi)$, $h_a(\cdot;\xi)$, $g_s(\cdot;\theta)$ and $h_s(\cdot;\zeta)$. In the second stage, we set $\lambda_i$ to $\{0.0018, 0.0035, 0.013, 0.025, 0.0483\}$ and train the LPM $P_E(\cdot;\psi)$ and $P_D(\cdot;\omega)$ separately for each Lagrangian operator. In both stages, we use Adam optimizer with batch size set to 12 and learning rate set to $10^{-4}$. The two phases are trained for 400 epochs and 50 epochs, respectively.

\subsection{Contributions of Prompt Tuning}
The role of additional prompts in compression is illustrated in Fig.~\ref{bitmap}. 
As the Lagrangian operator increases from left to right, the distortion measure in the rate-distortion loss function becomes a larger percentage. Consequently, the model increases bit consumption to minimize information loss in the image. In the third row, the attention map, generated by adding prompts in the STL, is visualized on the original image. As the $\lambda$ value increases, the model allocates more attention to the clock on the lighthouse and the lake in the lower left (darker red), and less attention to uninteresting regions such as the sky (darker blue). In the second row of the bit allocation map, the model assigns more bits (brighter yellow) to areas of interest like clocks and the lake. Generally, bit allocation rises as bit consumption increases, with highly focused areas receiving significantly more bits than less prominent ones. This bit allocation directly impacts the quality of the reconstructed image output by the model. As shown in the first row, the scale on the dial becomes clearer as the bit allocation increases, indicating the model's role in allocating more attention to this region.
\subsection{Comparison with Other Methods}
\noindent\textbf{Rate-distortion performance.} In Figure \ref{main experiment Kodak} and \ref{main experiment CLIC2020}, we present rate-distortion performance on the Kodak dataset~\cite{franzen1999kodak} and the CLIC2020 dataset~\cite{toderici2020clic}. Our model is compared with the state-of-the-art variable-rate method~\cite{gao2022flexible,cui2021asymmetric,song2021variable,yang2020variable,choi2019variable}, fixed-rate methods~\cite{lu2021transformer,mentzer2020high,xie2021enhanced,zou2022devil,cheng2020learned,minnen2018joint,balle2018variational} and traditional methods~\cite{bpg,bross2021overview}. PSNR and MS-SSIM are employed as an evaluation metric.

It can be obviously noticed that the proposed method outperforms all the current variable-rate methods. Also, we retrain the migratable method~\cite{gao2022flexible,cui2021asymmetric} with TIC~\cite{lu2021transformer} as the backbone to prevent the performance improvement due to different networks. In terms of fixed-rate methods, our approach compares favorably to the backbone TIC~\cite{lu2021transformer} and the current state-of-the-art image compression on PSNR and surpasses TIC~\cite{lu2021transformer} on MS-SSIM, while requiring far less storage space and training data than these methods, as demonstrated in the following sections.

\begin{figure}[h]
    \centering  
    \includegraphics[width=1\linewidth]{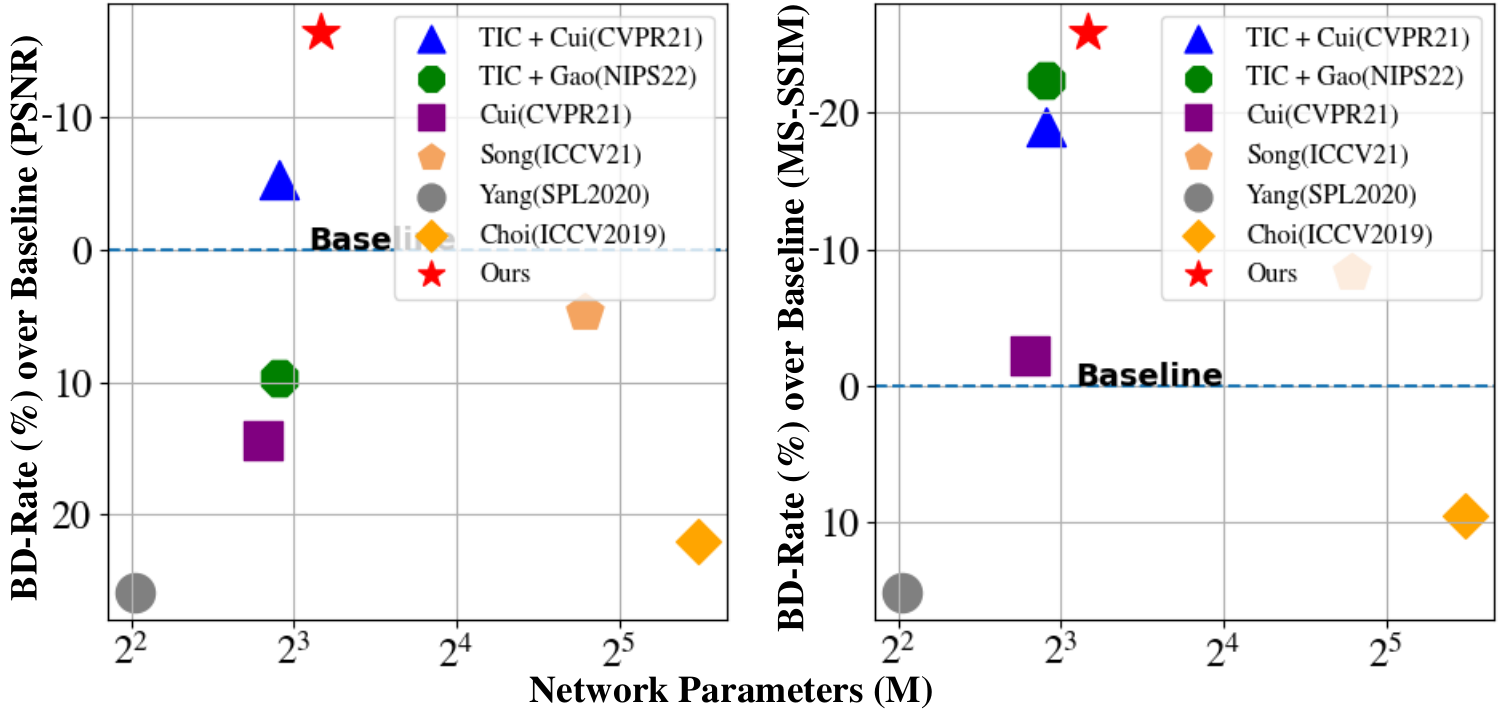}  
    \caption{Comparison of proposed method against other variable-Rate methods on Kodak dataset~\cite{franzen1999kodak} in terms of BD-rate (PSNR, left; MS-SSIM, right), where a smaller value indicates better performance. The baseline is BPG~\cite{bpg}.} 
    \label{model size}
    \vspace{-16pt}
\end{figure}

\noindent\textbf{Model size.} As shown in Figure \ref{model size}, we also compare the amount of storage with the variable-rate approach. Compared to methods~\cite{gao2022flexible,cui2021asymmetric,song2021variable,choi2019variable}, we achieve better performance with the same approximation or even less storage; whereas~\cite{yang2020variable} suffers from severe performance problems despite the fact that they require less storage.

\noindent\textbf{Qualitative results} Furthermore, we showcase visual comparisons of various methods~\cite{lu2021transformer,cui2021asymmetric,song2021variable,gao2022flexible}, as depicted in Figure \ref{visualization}. Compared to the base network TIC~\cite{lu2021transformer}, our method presents better visualization in the case of similar degree of compression. Compared to the variable-rate method~\cite{cui2021asymmetric,song2021variable,gao2022flexible}, we do not experience fuzzy block phenomenon despite consuming fewer bits.

\begin{table}[h]
\vspace{-8pt}
	\caption{BD-Rate results for different $\lambda$ on Kodak dataset~\cite{franzen1999kodak} and CLIC2020 dataset~\cite{toderici2020clic}. The baseline is TIC~\cite{lu2021transformer}.}
	\centering
	\resizebox{\columnwidth}{!}{
		\setlength{\tabcolsep}{.5em}{
			\begin{tabular}{crrrr}
				\toprule
				\multirow{2}{*}{Variant} & \multicolumn{2}{c}{\emph{Kodak~\cite{franzen1999kodak}} }                             & \multicolumn{2}{c}{\emph{CLIC2020~\cite{toderici2020clic}}}                          \\
				\cmidrule(l){2-3} \cmidrule(l){4-5} 
				& BD-Rate$_\text{P}$ & BD-Rate$_\text{M}$ & BD-Rate$_\text{P}$ & BD-Rate$_\text{M}$\\
				\midrule
				
				$\lambda_0$ & \textbf{0.54\%}  & \textbf{-2.33\%}  & \textbf{-0.17\%}  & \textbf{-4.66\%}  \\
				
				$\lambda_1$  & 11.55\% & -0.51\%  & 9.86\% & 2.47\% \\
				
				$\lambda_2$ & 4.08\% & -0.36\% & 0.51\% & -1.17\% \\
				
				$\lambda_3$ & 7.19\% & 6.60\% & 3.79\% & 6.13\% \\
				
				$\lambda_4$ & 22.63\% & 22.98\% & 18.19\% & 14.33\% \\
				
				$\lambda_5$ & 14.21\% & 10.47\% & 10.77\% & 4.28\% \\
				\bottomrule
	\end{tabular}}}
	\label{tab:variants}
\vspace{-8pt}
\end{table}

\subsection{The Selection of Initial $\lambda$}

A crucial step in the proposed method is how to select the pre-trained Lagrange multipliers $\lambda$. To solve this problem, we choose $\lambda_i$ as the initial Lagrange multiplier to train the backbone network separately. Subsequently, we adjust the target rate using different values and calculate the BD-Rate between each Rate-Distortion (R-D) curve and the TIC baseline. The results are as shown in Table \ref{tab:variants}.

It can be observed that the best performance is achieved when the $\lambda=0.0067$ pre-trained model is used as the backbone prompt-tuning, even surpassing the orignial network TIC~\cite{lu2021transformer} in terms of PSNR and MSSSIM performance for the CLIC dataset~\cite{toderici2020clic}.

\begin{figure}[h]

    \centering  
    \includegraphics[width=1\linewidth]{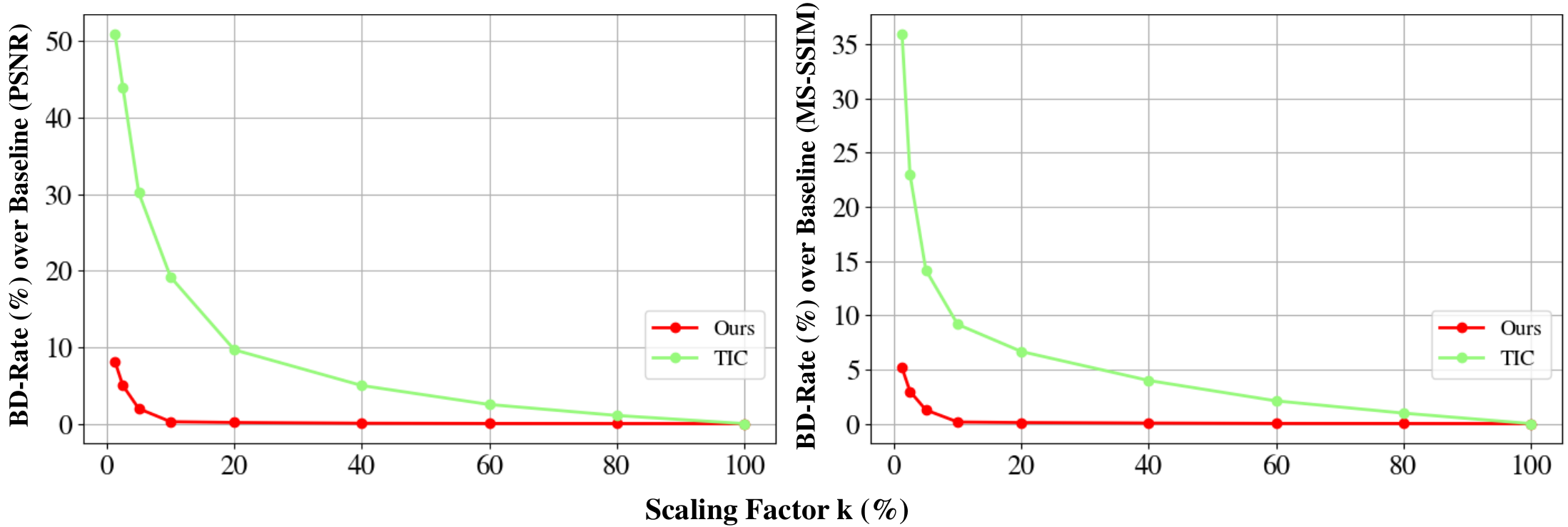}  
    \caption{Compare the performance degradation of TIC~\cite{lu2021transformer} and the proposed method on datasets with different scaling factors. The baseline is models trained respectively on the complete Flickr30k dataset~\cite{young2014image}.} 
    \label{sample}
    \vspace{-8pt}
\end{figure}

\subsection{Efficiency Comparison}
\noindent\textbf{Sample efficiency} We scale down the training data to different extents and then compare how it affects the performance of independently trained base networks and prompt-tuned base networks. Fig.~\ref{sample} shows that our method achieves satisfactory results with only 10\% of the original dataset, while the fixed-rate method experiences a significant decline in performance as the training data is reduced.

\begin{table}[h]
    \vspace{-8pt}
	\caption{The comparison of network parameter quantity and encoding/decoding speed.}
	\centering
	\resizebox{\columnwidth}{!}{
		\setlength{\tabcolsep}{.5em}{
			\begin{tabular}{ccccc}
				\toprule
				\multirow{2}{*}{Variant} & \multicolumn{2}{c}{Params}                             & \multicolumn{2}{c}{Time (s)}                          \\
				\cmidrule(l){2-3} \cmidrule(l){4-5} 
				& Encoder & Decoder & Encoding & Decoding\\
				\midrule
                LPM & 617.31K  & 848.45K & 0.004  & 0.009  \\

                TIC$_{ \text{w/o LPM}}$  & 3.65M & 3.86M  & 1.197 & 0.065 \\
                
				TIC$_{ \text{w LPM}}$ & 4.26M  & 4.68M & 1.201  & 0.074  \\
				\bottomrule
	\end{tabular}}}
	\label{storage}
    \vspace{-8pt}
\end{table}

\noindent\textbf{Storage and Training efficiency} Table \ref{storage} illustrates that LPM requires only 1.4M additional parameters, significantly less than the 7.51M needed for the standalone TIC~\cite{lu2021transformer}. When adjusting for a new target rate, it's sufficient to save just LPM, in contrast to the TIC, which necessitates storing a completely new model. This difference translates to a substantial 80\% reduction in storage space. Furthermore, incorporating LPM leads to a negligible increase of just 1\% in the time required for compression. During the training process, our method converges in just 50 epochs, significantly enhancing training efficiency compared to TIC~\cite{lu2021transformer}, which requires 400 epochs.

\section{Conclusion}
In this paper, we propose a variable bitrate fine-tuning method for transformer-based image compression.
 The Layer-adaptive Prompt Module (LPM) generates prompts to alter the model's attention allocation, thereby achieving variable bitrate compression. The proposed method achieves comparable performance with state-of-the-art fixed-rate methods and surpasses all existing variable-rate methods. Furthermore, it features efficient storage and sample utilization for resource-constrained environments.

\bibliographystyle{ieee_fullname}
\bibliography{egbib}

\begin{thebibliography}{10}\itemsep=-1pt

\bibitem{balle2016end}
Johannes Ball{\'e}, Valero Laparra, and Eero~P Simoncelli.
\newblock End-to-end optimized image compression.
\newblock {\em arXiv preprint arXiv:1611.01704}, 2016.

\bibitem{balle2018variational}
Johannes Ball{\'e}, David Minnen, Saurabh Singh, Sung~Jin Hwang, and Nick Johnston.
\newblock Variational image compression with a scale hyperprior.
\newblock {\em arXiv preprint arXiv:1802.01436}, 2018.

\bibitem{bpg}
Fabrice Bellard.
\newblock Bpg image format, 2018.

\bibitem{bross2021overview}
Benjamin Bross, Ye-Kui Wang, Yan Ye, Shan Liu, Jianle Chen, Gary~J Sullivan, and Jens-Rainer Ohm.
\newblock Overview of the versatile video coding (vvc) standard and its applications.
\newblock {\em IEEE Transactions on Circuits and Systems for Video Technology}, 31(10):3736--3764, 2021.

\bibitem{brown2020language}
Tom Brown, Benjamin Mann, Nick Ryder, Melanie Subbiah, Jared~D Kaplan, Prafulla Dhariwal, Arvind Neelakantan, Pranav Shyam, Girish Sastry, Amanda Askell, et~al.
\newblock Language models are few-shot learners.
\newblock {\em Advances in neural information processing systems}, 33:1877--1901, 2020.

\bibitem{chen2023transtic}
Yi-Hsin Chen, Ying-Chieh Weng, Chia-Hao Kao, Cheng Chien, Wei-Chen Chiu, and Wen-Hsiao Peng.
\newblock Transtic: Transferring transformer-based image compression from human perception to machine perception.
\newblock In {\em Proceedings of the IEEE/CVF International Conference on Computer Vision}, pages 23297--23307, 2023.

\bibitem{cheng2020learned}
Zhengxue Cheng, Heming Sun, Masaru Takeuchi, and Jiro Katto.
\newblock Learned image compression with discretized gaussian mixture likelihoods and attention modules.
\newblock In {\em Proceedings of the IEEE/CVF conference on computer vision and pattern recognition}, pages 7939--7948, 2020.

\bibitem{choi2019variable}
Yoojin Choi, Mostafa El-Khamy, and Jungwon Lee.
\newblock Variable rate deep image compression with a conditional autoencoder.
\newblock In {\em Proceedings of the IEEE/CVF International Conference on Computer Vision}, pages 3146--3154, 2019.

\bibitem{cui2021asymmetric}
Ze Cui, Jing Wang, Shangyin Gao, Tiansheng Guo, Yihui Feng, and Bo Bai.
\newblock Asymmetric gained deep image compression with continuous rate adaptation.
\newblock In {\em Proceedings of the IEEE/CVF Conference on Computer Vision and Pattern Recognition}, pages 10532--10541, 2021.

\bibitem{dosovitskiy2020image}
Alexey Dosovitskiy, Lucas Beyer, Alexander Kolesnikov, Dirk Weissenborn, Xiaohua Zhai, Thomas Unterthiner, Mostafa Dehghani, Matthias Minderer, Georg Heigold, Sylvain Gelly, et~al.
\newblock An image is worth 16x16 words: Transformers for image recognition at scale.
\newblock {\em arXiv preprint arXiv:2010.11929}, 2020.

\bibitem{franzen1999kodak}
Rich Franzen.
\newblock Kodak lossless true color image suite.
\newblock {\em source: http://r0k. us/graphics/kodak}, 4(2):9, 1999.

\bibitem{gao2022flexible}
Chenjian Gao, Tongda Xu, Dailan He, Yan Wang, and Hongwei Qin.
\newblock Flexible neural image compression via code editing.
\newblock {\em Advances in Neural Information Processing Systems}, 35:12184--12196, 2022.

\bibitem{jia2022visual}
Menglin Jia, Luming Tang, Bor-Chun Chen, Claire Cardie, Serge Belongie, Bharath Hariharan, and Ser-Nam Lim.
\newblock Visual prompt tuning.
\newblock In {\em European Conference on Computer Vision}, pages 709--727. Springer, 2022.

\bibitem{jin2023instance}
Feihu Jin, Jinliang Lu, Jiajun Zhang, and Chengqing Zong.
\newblock Instance-aware prompt learning for language understanding and generation.
\newblock {\em ACM Transactions on Asian and Low-Resource Language Information Processing}, 2023.

\bibitem{kim2022joint}
Jun-Hyuk Kim, Byeongho Heo, and Jong-Seok Lee.
\newblock Joint global and local hierarchical priors for learned image compression.
\newblock In {\em Proceedings of the IEEE/CVF Conference on Computer Vision and Pattern Recognition}, pages 5992--6001, 2022.

\bibitem{lee2018context}
Jooyoung Lee, Seunghyun Cho, and Seung-Kwon Beack.
\newblock Context-adaptive entropy model for end-to-end optimized image compression.
\newblock {\em arXiv preprint arXiv:1809.10452}, 2018.

\bibitem{liu2023pre}
Pengfei Liu, Weizhe Yuan, Jinlan Fu, Zhengbao Jiang, Hiroaki Hayashi, and Graham Neubig.
\newblock Pre-train, prompt, and predict: A systematic survey of prompting methods in natural language processing.
\newblock {\em ACM Computing Surveys}, 55(9):1--35, 2023.

\bibitem{lu2021transformer}
Ming Lu, Peiyao Guo, Huiqing Shi, Chuntong Cao, and Zhan Ma.
\newblock Transformer-based image compression.
\newblock {\em arXiv preprint arXiv:2111.06707}, 2021.

\bibitem{mentzer2020high}
Fabian Mentzer, George~D Toderici, Michael Tschannen, and Eirikur Agustsson.
\newblock High-fidelity generative image compression.
\newblock {\em Advances in Neural Information Processing Systems}, 33:11913--11924, 2020.

\bibitem{minnen2018joint}
David Minnen, Johannes Ball{\'e}, and George~D Toderici.
\newblock Joint autoregressive and hierarchical priors for learned image compression.
\newblock {\em Advances in neural information processing systems}, 31, 2018.

\bibitem{minnen2020channel}
David Minnen and Saurabh Singh.
\newblock Channel-wise autoregressive entropy models for learned image compression.
\newblock In {\em 2020 IEEE International Conference on Image Processing (ICIP)}, pages 3339--3343. IEEE, 2020.

\bibitem{qian2022entroformer}
Yichen Qian, Ming Lin, Xiuyu Sun, Zhiyu Tan, and Rong Jin.
\newblock Entroformer: A transformer-based entropy model for learned image compression.
\newblock {\em arXiv preprint arXiv:2202.05492}, 2022.

\bibitem{rabbani2002overview}
Majid Rabbani and Rajan Joshi.
\newblock An overview of the jpeg 2000 still image compression standard.
\newblock {\em Signal processing: Image communication}, 17(1):3--48, 2002.

\bibitem{shannon1959coding}
Claude~E Shannon et~al.
\newblock Coding theorems for a discrete source with a fidelity criterion.
\newblock {\em IRE Nat. Conv. Rec}, 4(142-163):1, 1959.

\bibitem{song2021variable}
Myungseo Song, Jinyoung Choi, and Bohyung Han.
\newblock Variable-rate deep image compression through spatially-adaptive feature transform.
\newblock In {\em Proceedings of the IEEE/CVF International Conference on Computer Vision}, pages 2380--2389, 2021.

\bibitem{toderici2015variable}
George Toderici, Sean~M O'Malley, Sung~Jin Hwang, Damien Vincent, David Minnen, Shumeet Baluja, Michele Covell, and Rahul Sukthankar.
\newblock Variable rate image compression with recurrent neural networks.
\newblock {\em arXiv preprint arXiv:1511.06085}, 2015.

\bibitem{toderici2020clic}
George Toderici, Lucas Theis, Nick Johnston, Eirikur Agustsson, Fabian Mentzer, Johannes Ball{\'e}, Wenzhe Shi, and Radu Timofte.
\newblock Clic 2020: Challenge on learned image compression, 2020, 2020.

\bibitem{toderici2017full}
George Toderici, Damien Vincent, Nick Johnston, Sung Jin~Hwang, David Minnen, Joel Shor, and Michele Covell.
\newblock Full resolution image compression with recurrent neural networks.
\newblock In {\em Proceedings of the IEEE conference on Computer Vision and Pattern Recognition}, pages 5306--5314, 2017.

\bibitem{xie2021enhanced}
Yueqi Xie, Ka~Leong Cheng, and Qifeng Chen.
\newblock Enhanced invertible encoding for learned image compression.
\newblock In {\em Proceedings of the 29th ACM international conference on multimedia}, pages 162--170, 2021.

\bibitem{yang2020variable}
Fei Yang, Luis Herranz, Joost Van De~Weijer, Jos{\'e} A~Iglesias Guiti{\'a}n, Antonio~M L{\'o}pez, and Mikhail~G Mozerov.
\newblock Variable rate deep image compression with modulated autoencoder.
\newblock {\em IEEE Signal Processing Letters}, 27:331--335, 2020.

\bibitem{young2014image}
Peter Young, Alice Lai, Micah Hodosh, and Julia Hockenmaier.
\newblock From image descriptions to visual denotations: New similarity metrics for semantic inference over event descriptions.
\newblock {\em Transactions of the Association for Computational Linguistics}, 2:67--78, 2014.

\bibitem{zhu2021transformer}
Yinhao Zhu, Yang Yang, and Taco Cohen.
\newblock Transformer-based transform coding.
\newblock In {\em International Conference on Learning Representations}, 2021.

\bibitem{zou2022devil}
Renjie Zou, Chunfeng Song, and Zhaoxiang Zhang.
\newblock The devil is in the details: Window-based attention for image compression.
\newblock In {\em Proceedings of the IEEE/CVF conference on computer vision and pattern recognition}, pages 17492--17501, 2022.

\end{thebibliography}
\clearpage
\addcontentsline{toc}{section}{Appendix}
\section*{Appendix}
\appendix
\renewcommand{\thetable}{A\arabic{table}}
\renewcommand{\thefigure}{A\arabic{figure}}

\end{document}